\documentclass[conference]{IEEEtran}
\IEEEoverridecommandlockouts
\usepackage{spconf,amsmath,epsfig}

\let\OLDthebibliography\thebibliography
\renewcommand\thebibliography[1]{
  \OLDthebibliography{#1}
  \setlength{\parskip}{0pt}
  \setlength{\itemsep}{0pt plus 0.3ex}
}
\usepackage{color}
\usepackage{booktabs}
\usepackage{arydshln}
\usepackage{multirow}
\usepackage{hyperref,MnSymbol}
\begin{document}\sloppy

% Example definitions.
% --------------------
\def\x{{\mathbf x}}
\def\L{{\cal L}}

% Title.
% ------
\title{EEP-3DQA: Efficient and Effective Projection-based 3D Model  \\ Quality Assessment}
%
% Single address.
% ---------------
% \name{Anonymous ICME submission}
\name{Zicheng Zhang$\tiny{^{1}}$,Wei Sun$\tiny{^{1}}$,Yingjie Zhou$\tiny{^{1}}$,Wei Lu$\tiny{^{1}}$,Yucheng Zhu$\tiny{^{1}}$,Xiongkuo Min,and Guangtao Zhai$\tiny{^{1,2}}$}
%Address and e-mail should NOT be added in the submission paper. They should be present only in the camera ready paper. 
\address{$^{1}$Institute of Image Communication and Network Engineering, Shanghai Jiao Tong University\\
 $^{2}$ MoE Key Lab of Artificial Intelligence, AI Institute, Shanghai Jiao Tong University
\thanks{This work was supported in part by NSFC (No.62225112, No.61831015), the Fundamental Research Funds for the Central Universities, National Key R\&D Program of China 2021YFE0206700, and Shanghai Municipal Science and Technology Major Project (2021SHZDZX0102).}}

\maketitle

\begin{abstract}
Currently, great numbers of efforts have been put into improving the effectiveness of 3D model quality assessment (3DQA) methods. However, little attention has been paid to the computational costs and inference time, which is also important for practical applications. Unlike 2D media, 3D models are represented by more complicated and irregular digital formats, such as point cloud and mesh. Thus it is normally difficult to perform an efficient module to extract quality-aware features of 3D models. In this paper, we address this problem from the aspect of projection-based 3DQA and develop a no-reference (NR) \underline{E}fficient and \underline{E}ffective \underline{P}rojection-based \underline{3D} Model \underline{Q}uality \underline{A}ssessment (\textbf{EEP-3DQA}) method. The input projection images of EEP-3DQA are randomly sampled from the six perpendicular viewpoints of the 3D model and are further spatially downsampled by the grid-mini patch sampling strategy. Further, the lightweight Swin-Transformer tiny is utilized as the backbone to extract the quality-aware features. Finally, the proposed EEP-3DQA and EEP-3DQA-t (tiny version) achieve the best performance than the existing state-of-the-art NR-3DQA methods and even outperforms most full-reference (FR) 3DQA methods on the point cloud and mesh quality assessment databases while consuming less inference time than the compared 3DQA methods. 
\end{abstract}
\section{Introduction}
\label{sec:intro}

3D models such as point cloud and mesh have been widely studied and applied in virtual/augmented reality (V/AR), game industry, film post-production, etc. However, the 3D models are usually bothered by geometry/color noise and compression/simplification loss during generation and transmission procedures. Therefore, many 3D model quality assessment (3DQA) methods have been proposed to predict the visual quality levels of degraded 3D models. Nevertheless, due to the complex structure of the 3D models, efficient feature extraction is difficult to perform and most 3DQA methods require huge computational resources and inference time, which makes it hard to put such methods into practical use and calls for more efficient 3DQA solutions. 

Normally speaking, 3DQA methods can be categorized into model-based and projection-based methods.
Different from model-based 3DQA methods that extract features directly for the 3D models, projection-based 3DQA methods evaluate the visual quality of 3D models via the 2D projections (regardless of the 3D models' digital representation formats and resolution), which can take advantage of the mature 2D vision backbones to achieve cost-effective performance. Unfortunately, the projections are highly dependent on the viewpoints and a single projection is not able to cover sufficient quality information. Therefore, many projection-based 3DQA methods try to utilize multiple projections or perceptually select the main viewpoint to achieve higher performance and gain more robustness \cite{liu2021pqa,zhang2022treating,nehme2022textured,zhang2022mm,fan2022no}. Namely, VQA-PC \cite{zhang2022treating} uses 120 projections for analysis and G-LPIPS \cite{nehme2022textured} needs to select the main viewpoint that covers the most geometric, color, and semantic information in advance. However, multiple projections and perceptual selection lead to taking up extra rendering time and huge computational resources, which motivates us to develop an \underline{E}fficient and \underline{E}ffective \underline{P}rojection-based \underline{3D} Model \underline{Q}uality \underline{A}ssessment (\textbf{EEP-3DQA}) method based on fewer projections.

Specifically, we propose a random projection sampling (RPS) strategy to sample projections from the 6 perpendicular viewpoints to reduce the rendering time cost. The tiny version \textbf{EEP-3DQA-t} only employs 2 projections while the base version \textbf{EEP-3DQA} employs 5 projections. The number of the projections is defined according to the experimental discussion in Section \ref{sec:num}. Then, inspired by \cite{wu2022fast}, we employ the Grid Mini-patch Sampling (GMS) strategy and adopt the lightweight Swin-Transformer tiny (ST-t) as the feature extraction backbone \cite{liu2021swin} since ST-t has a hierarchical structure and processes inputs with patch-wise operations, therefore it is naturally suitable for processing grid mini-patch maps.)  With the features extracted from the sampled projections, the fully-connected layers are used to map the features into quality scores. Finally, we average the quality scores of the sampled projections as the final quality value for the 3D model. The extensive experimental results show that the proposed method outperforms the existing NR-3DQA methods on the point cloud quality assessment (PCQA) and mesh quality assessment (MQA) databases, and is even superior to the FR-3DQA methods on the PCQA databases. Our proposed tiny version only takes about 1.67s to evaluate one point cloud on CPU (11.50$\times$ faster than VQA-PC) while still obtaining competitive performance.

\begin{figure}
    \centering
    \includegraphics[width = \linewidth]{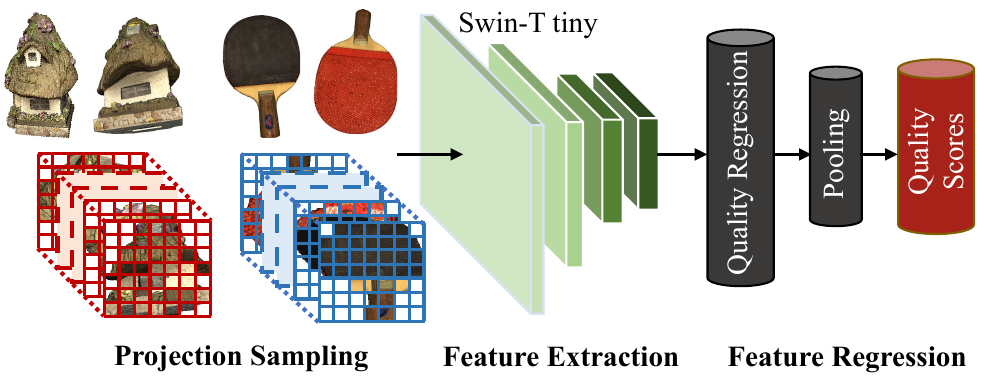}
    \caption{The framework of the proposed method.}
    \label{fig:framework}
    \vspace{-0.2cm}
\end{figure}

\section{Related Work}
In this section, we briefly review the development of model-based and projection-based 3DQA methods.
\subsection{Model-based 3DQA} 
The early FR-PCQA methods only use the geometry information to estimate the quality loss at the point level \cite{mekuria2016evaluation,tian2017geometric}. Further, to deal with the colored point clouds, both geometry and color information are incorporated for analysis by calculating the similarity of various quality domains \cite{meynet2020pcqm,yang2020graphsim,alexiou2020pointssim,database}. Later, 3D-NSS \cite{zhang2022no} is proposed by quantifying the distortions of both point clouds and meshes via some classic Natural Scene Statistics (NSS) distributions. ResSCNN \cite{liu2022point} proposes an end-to-end sparse convolutional neural network (CNN) to learn the quality representation of the point clouds. Unlike the 3D models used for classification and segmentation, the 3D models for quality assessment are usually denser and contain more points/vertices, thus making the feature extraction more complicated.

\subsection{Projection-based 3DQA}
Similar to the FR image quality assessment (IQA) methods, the early projection-based FR-3DQA methods compute the quality loss between the projections rendered from the reference and the distorted 3D models \cite{tian-color,yang2020predicting,zhang2022perceptual}. Namely, Tian $et$ $al.$ \cite{tian-color} introduces a global distance over texture image using Mean Squared Error (MSE) to quantify the effect of color information. Yang $et$ $al.$ \cite{yang2020predicting} uses both projected and depth images upon six faces of a cube for quality evaluation. 

The performance of the projection-based methods is further boosted by the development of deep learning networks. PQA-net \cite{liu2021pqa} designs a multi-task-based shallow network and extracts features from multi-view projections of the distorted point clouds. VQA-PC \cite{zhang2022treating} proposes to treat the point clouds as moving camera videos by capturing frames along the defined circular pathways, and utilize both 2D-CNN and 3D-CNN for spatial and temporal feature extraction respectively. G-LPIPS \cite{nehme2022textured} perceptually selects the main viewpoint of the textured meshes and assesses the quality of the main viewpoint projection with CNN.

\section{Proposed Method}
The framework of the proposed method is illustrated in Fig \ref{fig:framework}, which includes the projection sampling process, the feature extraction module, and the feature regression module.
\begin{figure*}
    \centering
    \includegraphics[width = 0.9\linewidth]{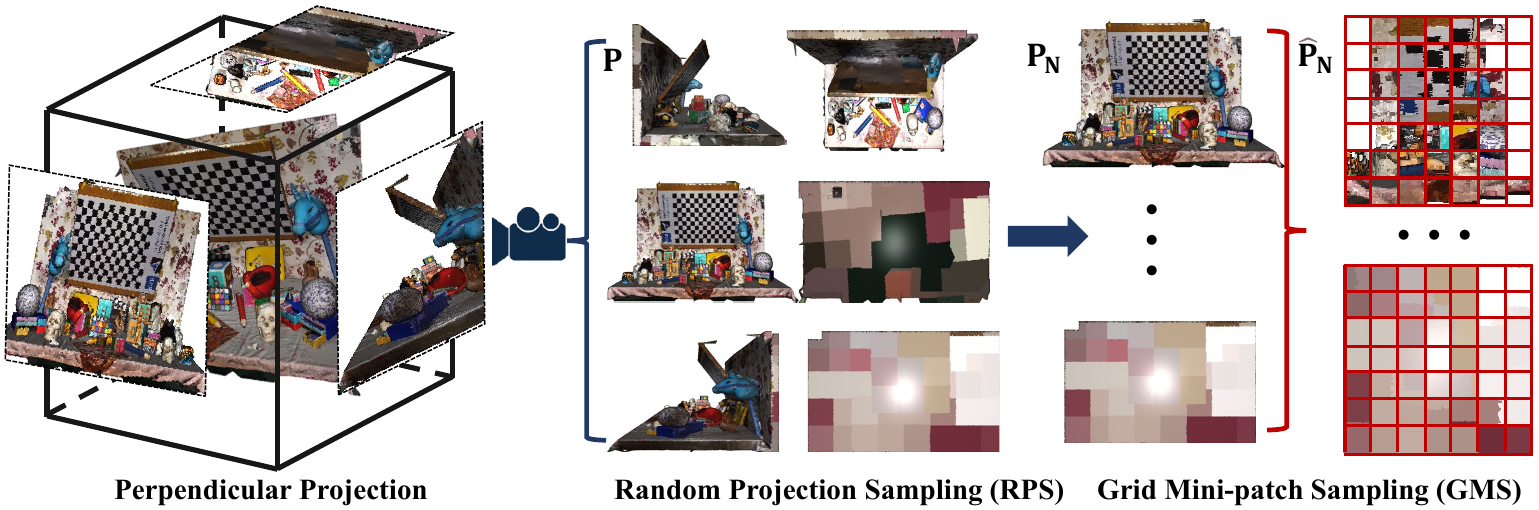}
    \caption{An example of the projection sampling process. $N$ projections are randomly selected from the 6 perpendicular viewpoints, which are further sampled into spatial grids by GMS. The details of GMS can be referred to in \cite{wu2022fast}.}
    \label{fig:grid}
\end{figure*}

\subsection{Projection Sampling Process}
\label{sec:projection}
Following the mainstream projection setting employed in the popular  point cloud compression standard MPEG VPCC \cite{graziosi2020overview}, we define 6 perpendicular viewpoints of the given 3D model $\mathbf{M}$ represented by point cloud or mesh, corresponding to the 6 surfaces of a cube:
\begin{equation}
\begin{aligned}
    \mathbf{P} &= \psi(\mathbf{M}), 
\end{aligned}
\end{equation}
where $\mathbf{P} = \{P_{i}|i=1,\cdots, 6\}$ indicates the set of the 6 projections and $\psi(\cdot)$ denotes the projection capture process. Additionally, the white background of the projections is cropped out.
There may exist redundant quality information among the 6 projections and the efficiency can be improved by extracting sufficient quality information from fewer projections since fewer projections consume less rendering time and computational resource. Therefore, we propose a Random Projection Sampling (RPS) strategy to improve the efficiency, which functions by randomly selecting $N$ projections for evaluation:
\begin{equation}
    \mathbf{P_N} = \alpha(\mathbf{P}),
\end{equation}
where $\alpha(\cdot)$ denotes for the RPS operation and $\mathbf{P_N} = \{P_{N_j}|j=1,\cdots, N\}$ stands for the set of sampled projections. It's worth noting that only the selected projections are rendered.

Further inspired by the boosted efficiency benefited from Grid Mini-patch Sampling (GMS) \cite{wu2022fast}, we similarly cut the projections into uniformly none-overlapped spliced spatial grids as the sampled results:
\begin{equation}
\begin{aligned}
    \mathbf{\hat{P}_N} &= \beta(\mathbf{P_N}),
\end{aligned}
\end{equation}
where $\beta$ indicates the GMS operation and $\mathbf{\hat{P}_N} = \{\hat{P}_{N_j}|j=1,\cdots, N\}$ represents the set of projections after GMS operation. From Fig. \ref{fig:grid}, we can see that the spatial grids can maintain the local quality-aware patterns that can be bothered with resize operation.

\subsection{Efficient Feature Extraction}
To take up fewer flops and parameters, we select the light-weight Swin-Transformer tiny (ST-t) \cite{liu2021swin} as the feature extraction backbone. Given the input projections set $\mathbf{\hat{P}_N}$, the quality-aware features can be obtained as:
\begin{equation}
\begin{aligned}
    F_{N_j} = &\gamma(\hat{P}_{N_j}), \\
    \overline{F}_{N_j} = &\mathbf{Avg}(F_{N_j}), \\
\end{aligned}
\end{equation}
where $\gamma(\cdot)$ represents the feature extraction operation with ST-t, $F_{N_j}$ indicates the extracted feature maps from the $N_j$-th input sampled projection, $\mathbf{Avg}(\cdot)$ stands for the average pooling operation and $\overline{F}_{N_j}$ denotes the pooled features.

\subsection{Quality Regression}
To map the quality-aware features into quality scores, we simply adopt a two-stage fully-connected (FC) layer for regression:
\begin{equation}
    Q_{N_j} = \mathbf{FC}(\overline{F}_{N_j}),
\end{equation}
where $Q_{N_j}$ indicates the quality score for the $N_j$-th sampled projection. Then the final quality $Q$ for the given 3D model can be computed by averaging the quality values:
\begin{equation}
    Q = \frac{1}{N} \sum_{j=1}^N  Q_{N_j},
\end{equation}
where $Q$ indicates the final quality score for the 3D model. The Mean Squared Error (MSE) is utilized as the loss function:
\begin{equation}
    Loss = \frac{1}{n} \sum_{\eta=1}^{n} (Q_{\eta}-Q_{\eta}')^2,
\end{equation}
where $Q_{\eta}$ is the predicted quality scores, $Q_{\eta}'$ is the quality label of the 3D model, and $n$ is the size of the mini-batch.

\begin{table*}[!htp]
\centering 
\caption{Performance results on the SJTU-PCQA and WPC databases. The best performance results are marked in {\bf\textcolor{red}{RED}} and the second performance results are marked in {\bf\textcolor{blue}{BLUE}}.}
% \vspace{-0.3cm}
\begin{tabular}{l:l:l|cccc|cccc}
\toprule
\multirow{2}{*}{Ref}&\multirow{2}{*}{Type}&\multirow{2}{*}{Methods} & \multicolumn{4}{c|}{SJTU-PCQA} & \multicolumn{4}{c}{WPC} \\ \cline{4-11}
        && & SRCC$\uparrow$      & PLCC$\uparrow$      & KRCC$\uparrow$     & RMSE $\downarrow$    & SRCC$\uparrow$      & PLCC$\uparrow$      & KRCC$\uparrow$       & RMSE $\downarrow$ \\ \hline
\multirow{10}{*}{FR} &\multirow{8}{45pt}{{{Model-based}}} 
 &MSE-p2po & 0.7294 & 0.8123 & 0.5617 & 1.3613 & 0.4558 & 0.4852 & 0.3182 & 19.8943 \\
 &&HD-p2po & 0.7157 & 0.7753 & 0.5447 & 1.4475 &0.2786 &0.3972&0.1943 &20.8990\\
 &&MSE-p2pl & 0.6277 & 0.5940 & 0.4825 & 2.2815 & 0.3281 & 0.2695 &0.2249 & 22.8226 \\
 &&HD-p2pl & 0.6441   & 0.6874    & 0.4565    & 2.1255 & 0.2827 & 0.2753 &0.1696 &21.9893 \\
 &&PSNR-yuv & 0.7950 & 0.8170 & 0.6196 & 1.3151 & 0.4493 & 0.5304 & 0.3198 & 19.3119\\
 && PCQM        & {0.8644}   & {0.8853}    & {0.7086}     & {1.0862}     &{0.7434}    & {0.7499}   & {0.5601}   & 15.1639             \\
&& GraphSIM    & {0.8783}    & 0.8449    & {0.6947}   & {1.0321}  & 0.5831    & 0.6163    & 0.4194   & 17.1939   \\
&& PointSSIM      & 0.6867  & 0.7136  & 0.4964 & 1.7001  & 0.4542    & 0.4667    & 0.3278   & 20.2733  \\ \cdashline{2-11}
&\multirow{2}{45pt}{{{Projection-based}}} &
PSNR &0.2952 &0.3222 &0.2048 &2.2972  &0.1261 &0.1801 &0.0897 &22.5482 \\
&&SSIM &0.3850 &0.4131 &0.2630 &2.2099  &0.2393 &0.2881 &0.1738 &21.9508 \\ \hdashline
\multirow{6}{*}{NR} &\multirow{2}{45pt}{{{Model-based}}} 
& 3D-NSS & 0.7144 & 0.7382  & 0.5174 & 1.7686    & 0.6479    & 0.6514    & 0.4417   & 16.5716 \\
&& ResSCNN & 0.8600 & 0.8100 &- &- &- &- &- &-\\ \cdashline{2-11}
&\multirow{4}{45pt}{{{Projection-based}}} &
PQA-net & 0.8500 & 0.8200 & - & - & 0.7000 & 0.6900 & 0.5100 & 15.1800 \\
&&VQA-PC & 0.8509 & 0.8635 & 0.6585 & 1.1334 & 0.7968 & 0.7976 & 0.6115 & 13.6219\\ \cdashline{3-11}
&&\textbf{EEP-3DQA-t} & \bf\textcolor{blue}{0.8891} & \bf\textcolor{blue}{0.9130} & \bf\textcolor{blue}{0.7324} & \bf\textcolor{blue}{0.9741} & \bf\textcolor{blue}{0.8032}	& \bf\textcolor{blue}{0.8124}	& \bf\textcolor{blue}{0.6176}	& \bf\textcolor{blue}{12.9603}
\\
&&\textbf{EEP-3DQA} &\bf\textcolor{red}{0.9095} &\bf\textcolor{red}{0.9363}	& \bf\textcolor{red}{0.7635}	& \bf\textcolor{red}{0.8472} & \bf\textcolor{red}{0.8264} & \bf\textcolor{red}{0.8296}	& \bf\textcolor{red}{0.6422}	& \bf\textcolor{red}{12.7451} 
 \\

\bottomrule
\end{tabular}
%\vspace{-0.4cm}
\label{tab:pcqa}
\end{table*}

\section{Experiment}

\subsection{Benchmark Databases}
To investigate the efficiency and effectiveness of the proposed method, the subjective point cloud assessment database (SJTU-PCQA) \cite{yang2020predicting}, the Waterloo point cloud assessment database (WPC) proposed by \cite{liu2022perceptual}, and the textured mesh quality (TMQ) database proposed by \cite{nehme2022textured} are selected for validation. The SJTU-PCQA database introduces 9 reference point clouds and each reference point cloud is degraded into 42 distorted point clouds, which generates 378 = 9$\times$7$\times$6 distorted point clouds in total.
The WPC database contains 20 reference point clouds and augmented each point cloud into 37 distorted stimuli, which generates 740 = 20$\times$37 distorted point clouds. The TMQ database includes 55 source textured meshes and 3,000 corrupted textured meshes with quality labels. The 9-fold cross validation is utilized for the SJTU-PCQA database and the 5-fold cross validation is used for the WPC and the TMQ databases respectively. The average performance is recorded as the final performance results.

\subsection{Competitors \& Criteria}
For the PCQA databases, the compared FR quality assessment methods include MSE-p2point (MSE-p2po) \cite{mekuria2016evaluation}, Hausdorff-p2point (HD-p2po) \cite{mekuria2016evaluation}, MSE-p2plane (MSE-p2pl) \cite{tian2017geometric}, Hausdorff-p2plane (HD-p2pl) \cite{tian2017geometric}, PSNR-yuv \cite{torlig2018novel}, PCQM \cite{meynet2020pcqm}, GraphSIM \cite{yang2020graphsim}, PointSSIM \cite{alexiou2020pointssim}, PSNR, and SSIM \cite{ssim}. The compared NR methods include 3D-NSS \cite{zhang2022no}, ResSCNN \cite{liu2022point}, PQA-net \cite{liu2021pqa}, and VQA-PC \cite{zhang2022treating}. For the TMQ database, the compared FR quality assessment methods include PSNR, SSIM \cite{ssim}, and G-LPIPS (specially designed for textured meshes) \cite{nehme2022textured}. The compared NR methods include 3D-NSS \cite{zhang2022no}, BRISQUE \cite{mittal2012brisque}, and NIQE \cite{mittal2012making}. It's worth mentioning that PSNR, SSIM, BRISQUE, and NIQE are calculated on all 6 projections and the average scores are recorded. 

Afterward, a five-parameter logistic function is applied to map the predicted scores to subjective ratings, as suggested by \cite{antkowiak2000final}. Four popular consistency evaluation criteria are selected to judge the correlation between the predicted scores and quality labels, which consist of Spearman Rank Correlation Coefficient (SRCC), Kendall’s Rank Correlation Coefficient (KRCC), Pearson Linear Correlation Coefficient (PLCC), and Root Mean Squared Error (RMSE). A well-performing model should
obtain values of SRCC, KRCC, and PLCC close to 1, and
the value of RMSE near 0.

\subsection{Implementation Details}
The employed Swin-Transformer tiny \cite{liu2021swin} backbone is initialized with the weights pretrained on the ImageNet database \cite{deng2009imagenet}. The RPS parameter $N$ discussed in Section \ref{sec:projection} is set as 2 for the tiny version \textbf{EPP-3DQA-t} and 5 for the base version \textbf{EPP-3DQA} respectively.
The adam optimizer \cite{kingma2014adam} is employed with the 1e-4 initial learning rate and the learning rate decays with a ratio of 0.9 for every 5 epochs. The default batch size is set as 32 and the default training epochs are set as 50. The average performance of $k$-fold cross validation is reported as the final performance to avoid randomness.

\begin{table}[!t]
\centering 
\renewcommand\tabcolsep{2.3pt}
\caption{Performance results on the TMQ database. The best performance results are marked in {\bf\textcolor{red}{RED}} and the second performance results are marked in {\bf\textcolor{blue}{BLUE}}.}
\vspace{0.1cm}
\begin{tabular}{l:l|cccc}
\toprule
\multirow{2}{*}{Ref}&\multirow{2}{*}{Methods} & \multicolumn{4}{c}{TMQ}  \\ \cline{3-6}
        & & SRCC$\uparrow$      & PLCC$\uparrow$      & KRCC$\uparrow$     & RMSE $\downarrow$  \\ \hline
\multirow{3}{*}{FR} &
PSNR &0.5295 &0.6535 &0.3938 &0.7877\\
&SSIM &0.4020 &0.5982 &0.2821 &0.8339 \\ 
&G-LPIPS &\bf\textcolor{red}{0.8600} & \bf\textcolor{red}{0.8500} &- &-\\ \hdashline
\multirow{5}{*}{NR}&
3D-NSS &0.4263	&0.4429	&0.2934	&1.0542  \\
&BRISQUE & 0.5364 & 0.4849  & 0.3788 & 0.9014    \\
&NIQE & 0.3731 & 0.3866  & 0.2528 & 0.8782    \\ \cdashline{2-6}
&\textbf{EEP-3DQA-t} & {0.7350} & {0.7430} & \bf\textcolor{blue}{0.5439} & \bf\textcolor{blue}{0.6468} 
\\
&\textbf{EEP-3DQA} &\bf\textcolor{blue}{0.7769} &\bf\textcolor{blue}{0.7823}	& \bf\textcolor{red}{0.5852}	& \bf\textcolor{red}{0.5975} 
 \\
\bottomrule
\end{tabular}
\vspace{-0.4cm}
\label{tab:mqa}
\end{table}

\subsection{Experimental Results}
The experimental results are listed in Table \ref{tab:pcqa} and Table \ref{tab:mqa}. From Table \ref{tab:pcqa}, we can observe that the proposed EEP-3DQA outperforms all the compared methods while the tiny version EEP-3DQA-t achieves second place on the SJTU-PCQA and WPC databases, which proves the effectiveness of the proposed method. What's more, all the 3DQA methods experience significant performance drops from the SJTU-PCQA database to the WPC database. This is because the WPC database introduces more complicated distortions and employs relatively fine-grained degradation levels, which makes the quality assessment tasks more challenging for the 3DQA methods. From Table \ref{tab:mqa}, we can find that the proposed EPP-3DQA is only inferior to the FR method G-LPIPS. However, G-LPIPS operates on the projections from the perceptually selected viewpoints, which thus gains less practical value than the proposed method.

\begin{figure}
    \centering
    \includegraphics[width=0.9\linewidth]{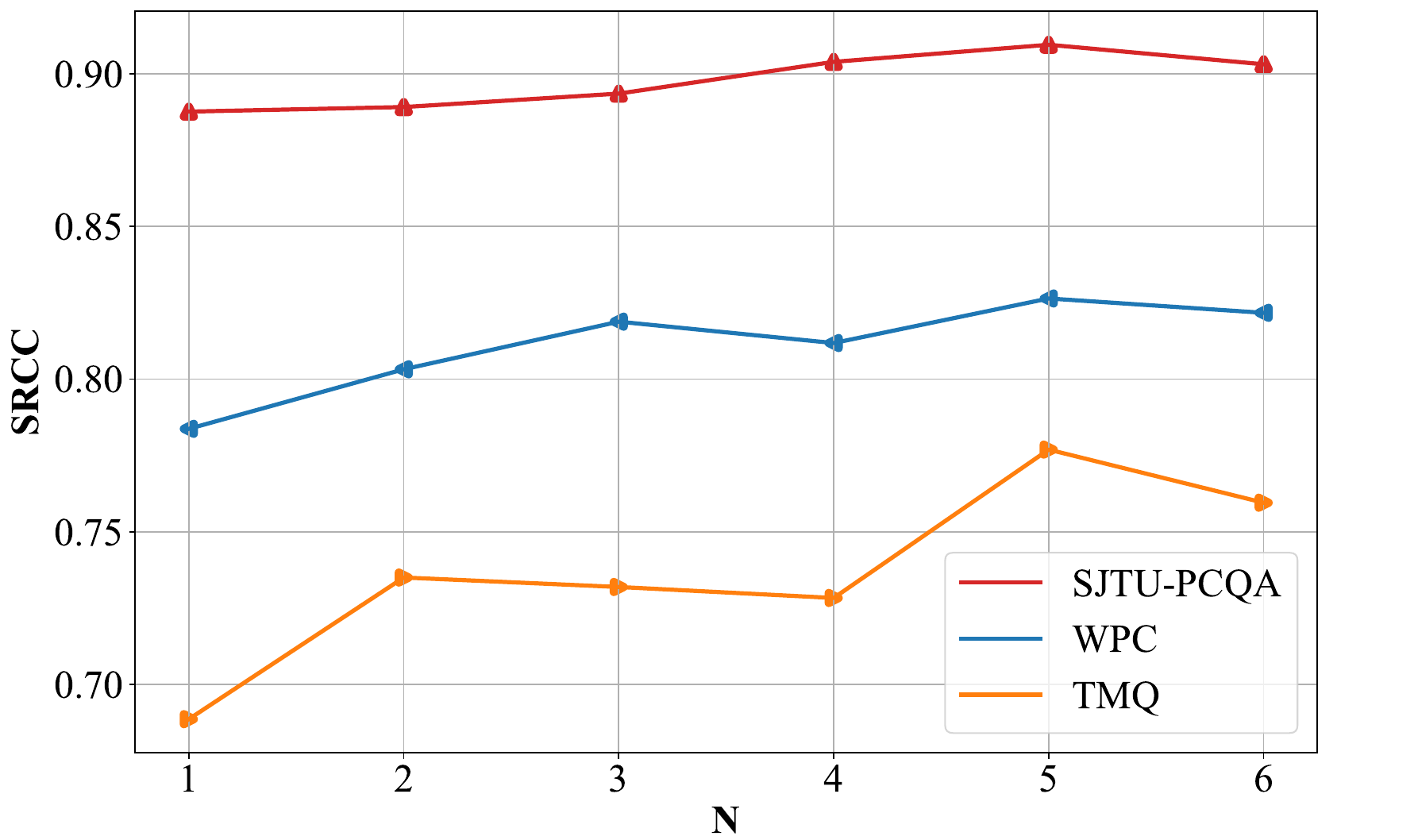}
    \vspace{-0.2cm}
    \caption{Illustration of the varying SRCC values corresponding to the different projection number $N$ selection. }
    \label{fig:n_srcc}
    \vspace{-0.4cm}
\end{figure}

\subsection{Projection Number Selection}
\label{sec:num}
In this section, we exhibit the performance with different RPS $N$ selections in Fig \ref{fig:n_srcc} and give the reasons for why setting $N$ as 2 for the tiny version and 5 for the base version of the proposed EPP-3DQA. From Fig \ref{fig:n_srcc}, we can see that randomly sampling 5 projections obtains the highest performance among all three databases, which is why we set $N$=5 for the base version. Interestingly, using all 6 projections causes performance drops. We attempt to give the reasons for such a phenomenon. Randomly selecting a smaller subset of projections could lead to better quality assessment performance than using all six projections, as it could reduce the influence of outliers and biases. Moreover, redundancy could be a possible reason why using all six projections may not necessarily lead to better quality assessment results. After experimenting with different numbers of projections, 5 projections seem to be the optimal choice for 3DQA tasks, especially for the TMQ database. Additionally, when $N$ increases from 1 to 2, the SRCC values gain the largest improvement on the SJTU-PCQA databases and gain relatively significant improvement on the WPC and TMQ databases. Therefore, we set $N$=2 for the tiny version rather than setting $N$=1 to get more cost-effective performance.

\subsection{Efficiency Analysis}
Previous discussions have proven the effectiveness of the proposed EEP-3DQA, this section mainly focuses on efficiency. Three FR-3DQA methods (PCQM, GraphSIM, and PointSSIM) and two NR-3DQA methods (3D-NSS and VQA-PC) are selected for comparison. It's worth mentioning that VQA-PC and the proposed EEP-3DQA are deep-learning based methods, and the rest methods are all handcrafted based methods. We test the operation time of the 3DQA methods on a computer with Intel 12500H @ 3.11 GHz CPU, 16G RAM, and NVIDIA Geforce RTX 3070Ti GPU on the Windows platform. The efficiency comparison is exhibited in Table \ref{tab:efficiency}. We can see that the base version of the proposed method requires only 1/2.09 inference compared with the fastest competitor 3D-NSS while achieving the best performance on CPU. Moreover, the tiny version EEP-3DQA-t takes up even fewer flops and inference time with the compared deep-learning based VQA-PC. All the comparisons confirm the superior efficiency of the proposed EPP-3DQA.

\subsection{Ablation Study}
We propose two sampling strategies RPS and GMS in Section \ref{sec:projection} and we try to investigate the contributions of each strategy. Note that we fix the projection viewpoints as not using the RPS strategy and we test with 5 different sets of fixed projection viewpoints to ease the effect of randomness. The ablation study results are shown in Table \ref{tab:ablation}, from which we can see that using both RPS and GMS achieves higher SRCC values than excluding either of the strategies. This indicates that the proposed RPS and GMS all make contributions to the final results. With closer inspections, we can find that RPS makes relatively more contributions to the tiny version compared with the base version. This is because the base version employs five random projections out of the six perpendicular projections, which is not that significantly different from fixing five projections out of the six perpendicular projections.  Additionally, the GMS strategy tends to make more contributions for the base version than the tiny version, which suggests that the GMS strategy can better dig out the quality-aware information with more projections.

\begin{table}[!tp]
    \centering
    \renewcommand\tabcolsep{2.3pt}
    \renewcommand\arraystretch{1.1}
    \caption{Illustration of flops, parameters, and average inference time (on CPU/GPU) per point cloud of the SJTU-PCQA and WPC databases. The subscript `$\mathbf{_{A\times}}$' of the consuming time indicates the corresponding method takes up $\mathbf{A\times}$ operation time of the proposed base version   \textbf{EEP-3DQA}. }
    \vspace{0.1cm}
    \begin{tabular}{c|cc|c}
    \toprule
    {Method} & {Para. (M)} & {Gflops} & {Time (S) CPU/GPU} \\ \hline
        PCQM      & - & -  & 12.23{$\mathbf{_{4.99\times}}$}/-  \\
        GraphSIM  & - & - &270.14{$\mathbf{_{110.26\times}}$}/-  \\ 
       
        PointSSIM &- & -   &9.27{$\mathbf{_{3.78\times}}$}/-  \\
        3D-NSS    &- &-    &5.12{$\mathbf{_{2.09\times}}$}/-   \\ 
        VQA-PC    &58.37 &50.08   &19.21{$\mathbf{_{7.84\times}}$}/16.44{$\mathbf{_{11.26\times}}$}   \\ \hdashline
        \textbf{EEP-3DQA-t} &27.54 & 8.74  &1.67\textcolor{red}{$\mathbf{_{0.68\times}}$}/1.12\textcolor{red}{$\mathbf{_{0.77\times}}$} \\ 
        \textbf{EEP-3DQA}   &27.54 &21.87    &2.45\textcolor{blue}{$\mathbf{_{1.00\times}}$}/1.46\textcolor{blue}{$\mathbf{_{1.00\times}}$} \\ 
    \bottomrule
    \end{tabular}
    %\vspace{-0.4cm}
    \label{tab:efficiency}
\end{table}

\begin{table}[]
\centering
\renewcommand\arraystretch{1.1}
\caption{SRCC performance results of the ablation study. Best in bold.}
\vspace{0.05cm}
\begin{tabular}{c:c:c|c|c|c}
\toprule
Ver. & RPS & GMS & SJTU & WPC &TMQ \\ \hline
\multirow{4}{*}{Tiny} & $\times$ & $\times$ & 0.8711 & 0.7922 & 0.7255 \\
& $\checkmark$ & $\times$ & 0.8834 & 0.8016 & 0.7347 \\
& $\times$ & $\checkmark$ & 0.8822 & 0.7953 & 0.7136 \\
& $\checkmark$ & $\checkmark$ & \textbf{0.8891} & \textbf{0.8032} & \textbf{0.7350}  \\ \hdashline
\multirow{4}{*}{Base} & $\times$ & $\times$ & 0.8741 & 0.8012 & 0.7333 \\
& $\checkmark$ & $\times$ & 0.8853 & 0.8140 & 0.7564 \\
& $\times$ & $\checkmark$ & 0.8933 & 0.8241 & 0.7704 \\
& $\checkmark$ & $\checkmark$ & \textbf{0.9095} & \textbf{0.8264} & \textbf{0.7769} \\

\bottomrule
\end{tabular}
\vspace{-0.3cm}
\label{tab:ablation}
\end{table}

% \subsection{Cross Database Performance}
\section{Conclusion}
In this paper, we mainly focus on the efficiency of 3DQA methods and propose an NR-3DQA method to tackle the challenges. The proposed EEP-3DQA first randomly samples several projections from the 6 perpendicular viewpoints and then employs the grid mini-patch sampling to convert the projections into spatial grids while marinating the local patterns. Later, the lightweight Swin-Transformer tiny is used as the feature extraction backbone to extract quality-aware features from the sampled projections. The base EEP-3DQA achieves the best performance among the NR-3DQA methods on all three benchmark databases and the tiny EEP-3DQA-t takes up the least inference time on both CPU and GPU while still obtaining competitive performance. The further extensive experiment results further confirm the contributions of the proposed strategies and the rationality of the structure.

\bibliographystyle{IEEEbib}
\bibliography{icme2022template}

\end{document}